\documentclass[11pt, a4paper, logo, copyright]{googledeepmind}

\usepackage{graphicx} 
\usepackage[authoryear, compress, round]{natbib}
\usepackage{wrapfig}
\usepackage{xspace}
\usepackage{epigraph}
\usepackage{hyperref}
\usepackage[capitalize]{cleveref}
\usepackage{indentfirst}
\usepackage{enumitem}
\usepackage{tabularx}
\usepackage{makecell}
\usepackage{amsmath,amssymb}
\usepackage{amsthm}
\usepackage[linesnumbered,ruled,vlined]{algorithm2e}
\usepackage{subcaption}
\usepackage{bm}

\crefname{figure}{Figure}{Figures}
\newcommand{\R}{\mathbb{R}}

\title{Quantifying the Necessity of Chain of Thought through Opaque Serial Depth}

\correspondingauthor{jonahbc@google.com}

\renewcommand{\today}{}

\author[1]{Jonah Brown-Cohen}
\author[1]{David Lindner}
\author[1]{Rohin Shah}
\affil[1]{Google DeepMind}

\usepackage{tcolorbox}
\newtcolorbox{formattedquote}{
    colback=blue!3!white,
    colframe=blue!20!white,
    fontupper=\footnotesize,
    boxsep=-2pt 
}

\begin{abstract}
    Large language models (LLMs) tend to externalize their reasoning in their chain of thought, making the chain of thought a good target for monitoring. This is partially an inherent feature of the Transformer architecture: sufficiently long serial cognition must pass through the chain of thought~\citep{korbak2025chain}. We formalize this argument through the notion of \emph{opaque serial depth}, given by the length of the longest computation that can be done without the use of interpretable intermediate steps like chain of thought. Given this formalization, we compute numeric upper bounds on the opaque serial depth of Gemma 3 models, as well as asymptotic results for additional architectures beyond standard LLMs. We also open-source an automated method that can calculate upper bounds on the opaque serial depth of arbitrary neural networks, and use it to demonstrate that Mixture-of-Experts models likely have lower depth than dense models. Overall, our results suggest that opaque serial depth is a useful tool for understanding the potential for models to do significant reasoning that is not externalized.
\end{abstract}

\begin{document}

\maketitle

\section{Introduction}

\begin{figure}[ht] 
  \centering %
  \includegraphics[trim=12 12 108 12, clip, width=0.75\textwidth]{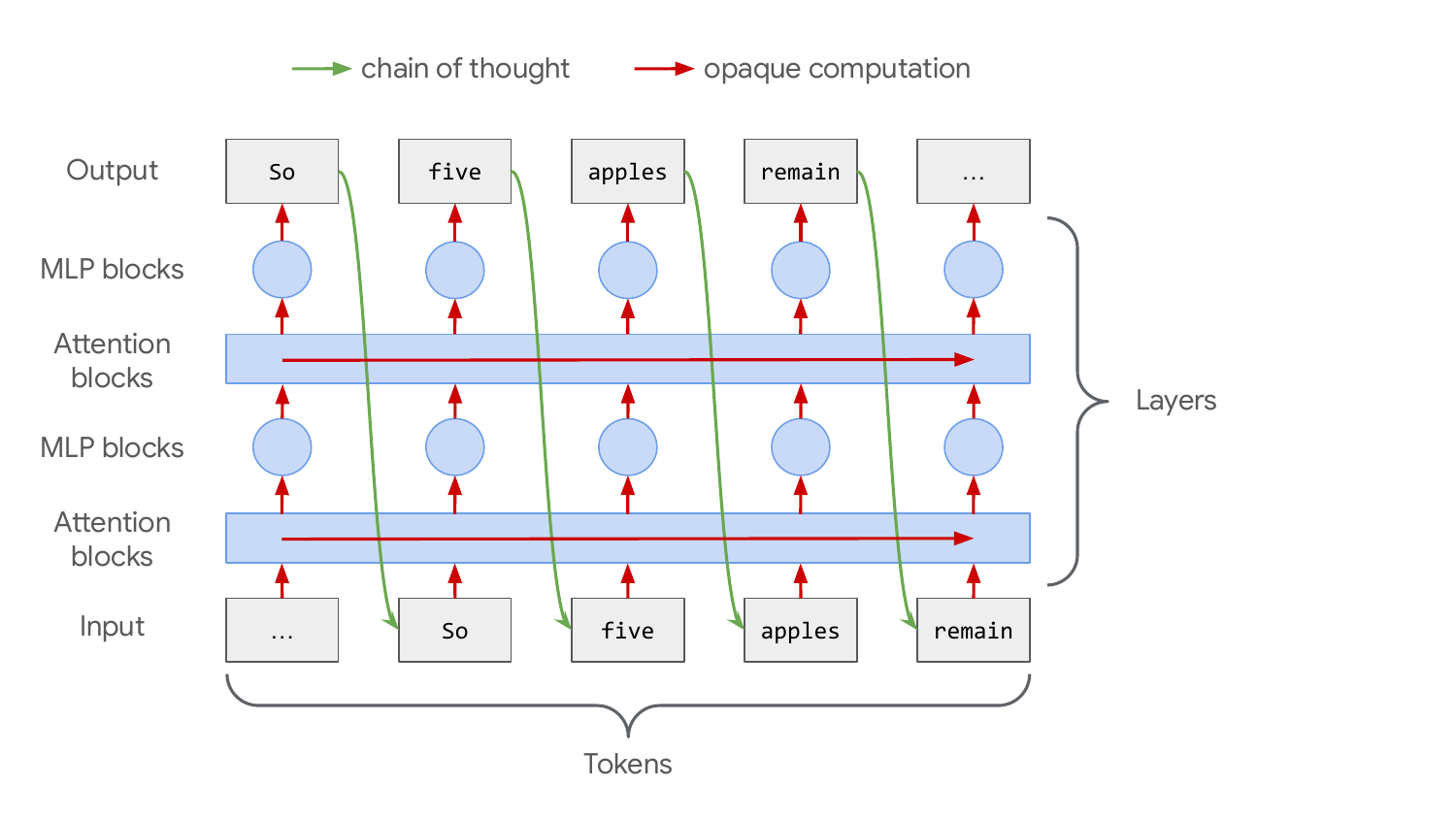}
  \caption{\emph{Adapted from \citet{korbak2025chain}.} For Transformers, chain of thought is the only way to pass information from later layers to earlier layers, making it a bottleneck for serial computation. As a result, for tasks that require sufficiently long serial computation, the model will have to externalize some of its reasoning in the chain of thought.} 
  \label{fig:cot-intuition}
\end{figure}

Chain of thought monitoring for large language models (LLMs) is an important mitigation for AI safety. A key intuition supporting its robustness is that ``thinking out loud is necessary for hard tasks''~\citep{korbak2025chain}, because there is no other way for LLMs based on the Transformer architecture~\citep{vaswani2017attention} to perform long serial cognition, as illustrated in \cref{fig:cot-intuition}. 

Different architectures could change these considerations. For example, adding recurrence should enable long serial cognition within the activations, and so should reduce our expectation that thinking out loud is necessary. As a result, we may want to avoid recurrent architectures, so as to preserve our ability to understand how an LLM is reasoning simply by reading its chain of thought.

To make such arguments more precisely and rigorously, it is useful to have a standard, well-founded notion of \emph{opaque serial depth}, that is, the extent to which the architecture supports long chains of cognition without having to think out loud. One approach would be to simply count the number of layers, but this immediately raises the question of what constitutes a layer. Should normalization operations be considered their own separate layer, or are they simply a part of an existing layer? Shouldn't a multi-headed attention layer be ``deeper'' than a linear layer?

Taking inspiration from computational complexity theory, we propose measuring the depth of a neural network based on the complexity measure known as circuit depth – the minimum depth of a Boolean circuit that computes the same function as the neural network. \cref{sec:why-circuit-depth} discusses the many advantageous properties that circuit depth enjoys, while \cref{sec:formalizing-depth} provides a formal definition. We demonstrate how to compute upper bounds on opaque serial depth in practice in \cref{sec:calculating-by-hand}, including by-hand calculations for models in the Gemma 3 family~\citep{team2025gemma} and asymptotic analysis for a few different architectures. In \cref{sec:automated-calculation}, we implement and open-source an automatic method for calculating upper bounds on opaque serial depth of arbitrary neural networks, and demonstrate that the resulting bounds are close to the tighter bounds calculated by hand. \cref{sec:limitations} discusses limitations and \cref{sec:conclusion} concludes.

\section{Why focus on circuit depth?} \label{sec:why-circuit-depth}

The notion of depth has been formalized and extensively studied in computational complexity theory, via the complexity measure known as \emph{circuit depth}. The circuit depth of a function (such as that computed by a neural network) is the minimum depth of a Boolean circuit that computes that function. This is a natural and useful choice to formalize intuitions around the necessity of chain of thought for hard tasks for a variety of reasons.

\paragraph{Implementation independence.} Since circuit depth is defined for a function, it does not depend on the particular implementation of the mathematical operations in a neural network, only on its expressive power. This allows us to sidestep questions about e.g. what should constitute an individual layer of a neural network.

\paragraph{Theoretical connections between circuit depth and chain of thought.} Recent work has used Boolean circuit depth to analyze the power and limitations of the standard Transformer architecture~\citep{vaswani2017attention}, particularly in explaining the power of chain of thought reasoning~\citep{hahn2020theoretical,merrill2022saturated,liu2024Chain}.
This line of work demonstrates that bounded-depth Transformers can be computed by bounded-depth Boolean circuits, and hence have strong limitations on what problems they can solve without chain of thought. These results are asymptotic: they apply to input sequences of length $n$ and give bounds on the circuit depth of $O(1)$ or $O(\log n)$ depending on the assumptions made.

\paragraph{Circuit depth distinguishes between parallel and serial reasoning.} If a given function $f$ can be computed by a depth-$d$ circuit, then one can compute $f$ in parallel in wall-clock time $O(d)$, using one processor for each gate in the circuit. Thus, the circuit depth of $f$ is a quantification of the minimum serial computation needed to compute $f$, while allowing vast parallel reasoning. This exactly mirrors the intuition for why chain of thought is necessary: while the Transformer architecture certainly does vast parallel computation on a sequence of tokens, it can only do limited serial computation, \emph{except} when using chain of thought. So, chain of thought should be necessary for tasks that benefit from significant serial computation (i.e. ``hard tasks'').

Computational complexity theory has already used circuit depth to demonstrate that serial reasoning can be a bottleneck for some tasks. Under widely believed conjectures, there are problems solvable in polynomial time that cannot be solved by low-depth circuits e.g. $O(\log n)$ depth, including, notably, the problem of planning in MDPs~\citep{papadimitriou1987complexity}. In our setting of interest, this would suggest that a neural network with limited depth would struggle to do a lot of planning within a single forward pass. We may also expect that the transparency and interpretability of the chain of thought would be better if the serial depth between successive tokens is smaller.

\section{Formalizing Opaque Serial Depth} \label{sec:formalizing-depth}

\subsection{Circuit Depth in Complexity Theory}
We begin by introducing the complexity-theoretic model of circuits with which we will measure depth.
There are various technical subtleties involving the number of bits of precision when computing mathematical operations with Boolean logical circuits. To avoid some of these complexities, we will assume that all mathematical operations are performed on floating-point numbers with the same, fixed number of bits of precision. This will allow us to present a somewhat simplified definition and method for calculating neural network depth. However, a neural network architecture using floating-point numbers with many more bits of precision than is standard must be evaluated differently.

\paragraph{Circuits.}
A \emph{circuit} is a directed acyclic graph that breaks down the computation of a given function into simpler operations. The edges in the graph are referred to as wires, and the nodes are referred to as gates. There are a set of $n$ designated \emph{input wires}, and a set of $m$ designated \emph{output gates}. For our purposes, each wire will carry a single real number, and each gate will compute a real-valued function applied to the real numbers carried on the wires feeding into that gate. The output of a circuit on a given real $n$-dimensional input $x$ is computed as follows:

\begin{enumerate}
\item Write the $n$ coordinates of $x$ on the $n$ input wires.
\item For each gate G that has a number written on all of its incoming wires:
\begin{enumerate}
\item Compute the real-valued function given by G.
\item Write the output value of G on all of its outgoing wires
\end{enumerate}
\item The m-dimensional output y is the vector with coordinates given by the values computed by the m output gates.
\end{enumerate}

\paragraph{Circuit Depth.}
To measure circuit depth meaningfully, we must restrict the mathematical operations allowed at each gate. This captures the intuition that a circuit is a way of specifying how to compute a more complex function via simple operations. The permissible gate operations we choose are:
\begin{enumerate}
\item Associative binary operations applied to two real numbers e.g. addition, multiplication, max, or min.
\item Piecewise functions of a single real number, where each piece is an analytic function.
\end{enumerate}
Given this set of gate operations the \emph{depth of a circuit} is given by the \emph{maximum length} of a path from an input wire to an output gate.

The reason to consider piecewise analytic functions is that there are known to be logarithmic depth Boolean circuits for powering, addition and multiplication \citep{Beame1984LogDC}. Thus for a fixed number of bits of precision $B$, we can approximate analytic functions by their Taylor series to sufficient accuracy using depth $O(\log B)$. For simplicity we absorb the constant depending on the bit-precision, and just model such operations as requiring depth 1.

\subsection{Serial Depth of a Neural Network} \label{sec:serial-depth-nn}

A neural network immediately corresponds to a function $f_{\theta}:\R^n \to \R^m$, where $\theta$ represents the weights of the neural network. We let $S$ be the number of parameters defining $\theta$.
We define the \emph{depth of a neural network} with weights $\theta$ as the the \emph{minimum depth} of a circuit of size \emph{polynomial in $S$} that computes the function $f_{\theta}$.
Overall this means that the depth of a neural network with weights $\theta$ is given by
\begin{equation}
\text{Depth}(f_\theta) = \min_{\text{poly}(S)\text{-size circuits } C}\max_{\text{ paths }P\text{ in } C} \text{Length}(P).
\end{equation}
The reason for the constraint that the circuit size be polynomial in $S$, is that one can always produce a depth $n$, size $2^{O(n)}$ circuit for any function $f$, by simply writing out a giant lookup table of the correct output for every possible input. This constraint is not particularly binding for current Transformer models, where the input dimension $n$ is the product of the sequence length and the embedding dimension, and hence is typically much larger than the depth of the natural circuit formed from the neural network's computational graph. However, future architectures could in principle have depth much larger than $n$, but a similar number of parameters $S$, and so it is important to include the above constraint.

In general, it is likely to be intractable to find the minimally deep circuit equivalent to a given neural network. However, it is straightforward to compute an upper bound on the neural network depth by finding \emph{any} circuit that computes $f_{\theta}$ and calculating its depth. This is the approach we take in this paper.

\subsection{Opaque Serial Depth of a Language Model}

The prior section provides a mathematically formal definition of neural network depth. However, when attempting to measure the ability of a neural network to perform opaque serial reasoning, there are additional considerations. For example, an autoregressive transformer model outputs one token at a time, then feeds the full sequence produced so far back in as input before producing the next output token. Overall we view the output of the model on a given prompt as the full sequence of output tokens. We would like our measurement of depth to capture the fact that while the internal computations of a neural network are not easily interpretable, intermediate steps in the chain of thought producing the output may form \emph{interpretable bottlenecks}. In the case of a transformer model, after each forward pass, a new token is produced, and the corresponding piece of natural language text represented by the token could reasonably be considered to be interpretable.
Similarly a text diffusion model produces intermediate token outputs on each diffusion step, and each of these intermediate steps could also be considered interpretable.

Motivated by these observations, we would like a definition of opaque serial depth that allows us to measure only the serial depth present between interpretable bottlenecks, such as intermediate tokens produced by transformers or text diffusion models.
To do this, we view any computation involving feeding outputs of a neural network into inputs of a (possibly the same) neural network as a single large circuit defined by connecting the output wires of some neural networks to the input wires of others.
For example, an autoregressive transformer applied to a prompt of length $k$ and fixed output length $n$ corresponds to a circuit with $n-k$ copies of the original transformer. The $i$-th copy of the circuit has input wires connected to the $k$ token prefix as well as the output wires of the $i-1$ previous copies. 

It is then straightforward to adapt the definition of depth to this setting by marking the outputs of certain gates/nodes in the circuit as \emph{interpretable}, including the overall inputs and outputs. In this case, one can perform a depth-first search from any interpretable node $u$ that terminates at a leaf whenever it reaches another interpretable node $v$. Thus, each interpretable node $u$ can be written as a function $f^{u}_{\theta}(x)$, where each coordinate of $x$ corresponds to some interpretable node $v$. The \emph{opaque serial depth} is then the maximum over interpretable nodes $u$ of $\text{Depth}(f^{u}_{\theta})$.

\subsection{What Counts as Interpretable?} \label{sec:interpretable-nodes}

The definition of opaque serial depth requires a user-specified set of ``interpretable'' nodes in a computational graph. Clearly, the decision of what to include in or exclude from this set will have a large impact on the calculated serial depth of a computation. Unfortunately, there is not yet a precise technical definition of interpretability that we could use to make this determination. Instead, we suggest two different methods for assessing whether a node is interpretable.

The first method draws on the intuition that if we ``understand'' something, then we should be able to answer questions about it. This suggests a simple test: for a node to be considered interpretable, we should be able to answer questions about the model's reasoning simply by looking at the information contained in that node. Monitorability evaluations~\citep{guan2025monitoring} are one way in which this can be done. We can get even stronger evidence that the node is interpretable through stress tests that actively try to find questions that can't be answered~\citep{emmons2025chain, deng2025cot}.

The second method is to reason about what the node was optimized to do during training. If a node was directly optimized to imitate human-written text (e.g. via pre-training or supervised fine-tuning) or to look good to humans (e.g. via RLHF), this gives it a ``natural language prior'' and we should expect the information in that node to be interpretable to humans. If there are other incentives during training, this argument becomes weaker. For example, reasoning training~\citep{jaech2024openai} provides a separate incentive for the chain of thought to be useful for solving difficult problems. So far reasoning training shapes the chain of thought less than pretraining, and so the chain of thought is likely still quite interpretable, but if it is scaled up significantly this may change.

For a significantly more detailed discussion of these two approaches, see \cref{appendix:interpretable-nodes}.

\section{Calculating Serial Depth} \label{sec:calculating-by-hand}

In this section we explain in detail how to calculate upper bounds on the serial depth of a neural network.

\subsection{Upper Bounds on the Depth of a Neural Network}
\label{sec:hand-upper-bound-alg}
Our definition of neural network depth refers to the minimum depth of any circuit computing the same function as the neural network. This is famously difficult to calculate due to the quantification over all possible circuits computing the function. However, there is a straightforward method to calculate an \emph{upper bound} on the circuit depth, by simply exhibiting a circuit of a given depth that correctly computes the function.

Any neural network implementation is already a real-valued circuit, though the gates can have more than two input wires. Thus, we need only account for the added depth required to compute operations on more than two inputs to determine the depth of this circuit, which then forms an upper bound for the opaque serial depth of the overall neural network.

These considerations lead to the following algorithm for computing an upper bound on the depth of a neural network:

\begin{algorithm}[H]
\caption{Opaque Serial Depth Calculation} \label{alg:opaque-serial-depth}
\DontPrintSemicolon

\SetKwFunction{FDepth}{Depth}
\SetKwFunction{FOpaque}{OpaqueSerialDepth}
\SetKwFunction{FImm}{Immediate\_Depth}
\SetKwProg{Fn}{Function}{:}{}

\Fn{\FOpaque{$node$, $I$}}{
    \tcp{Where $I$ is the interpretable\_nodes\_set}
    \KwRet $\max \left( \FDepth{node, I}, \; \max_{n \in I} \FDepth{n, I} \right)$\;
}
\BlankLine

\Fn{\FDepth{$node$, $I$}}{
    \If{$node \in I$}{
        \KwRet $0$\;
    }
    \KwRet $\FImm{node} + \max_{c \in children(node)} \FDepth{c, I}$\;
}

\Fn{\FImm{$node$}}{
    \If{$node$ is an associative binary operation on $n$ inputs}{
        \KwRet $\log_2 n$\;
    }
    \ElseIf{$node$ is a piecewise analytic function of two or fewer inputs}{
        \KwRet $1$\;
    }
}
\end{algorithm}

The algorithm performs a depth first search starting from each interpretable node, and recursively adds up the total opaque serial depth by computing the depth contributed by the current node plus the maximum depth of any child node. The depth first search terminates whenever it reaches an interpretable input nodes. When computing the depth of a circuit, our definition requires operations that take at most two inputs. Hence, associative binary operations taking $n > 2$ inputs must be computed by a depth $\log_2 n$ tree, where each node in the tree applies the binary associative operation on two inputs from the next level of the tree.

\subsection{Example: Two layer MLP}

\begin{figure}[th] 
  \centering %
  \includegraphics[trim=0 0 36 0, clip, width=0.9\textwidth]{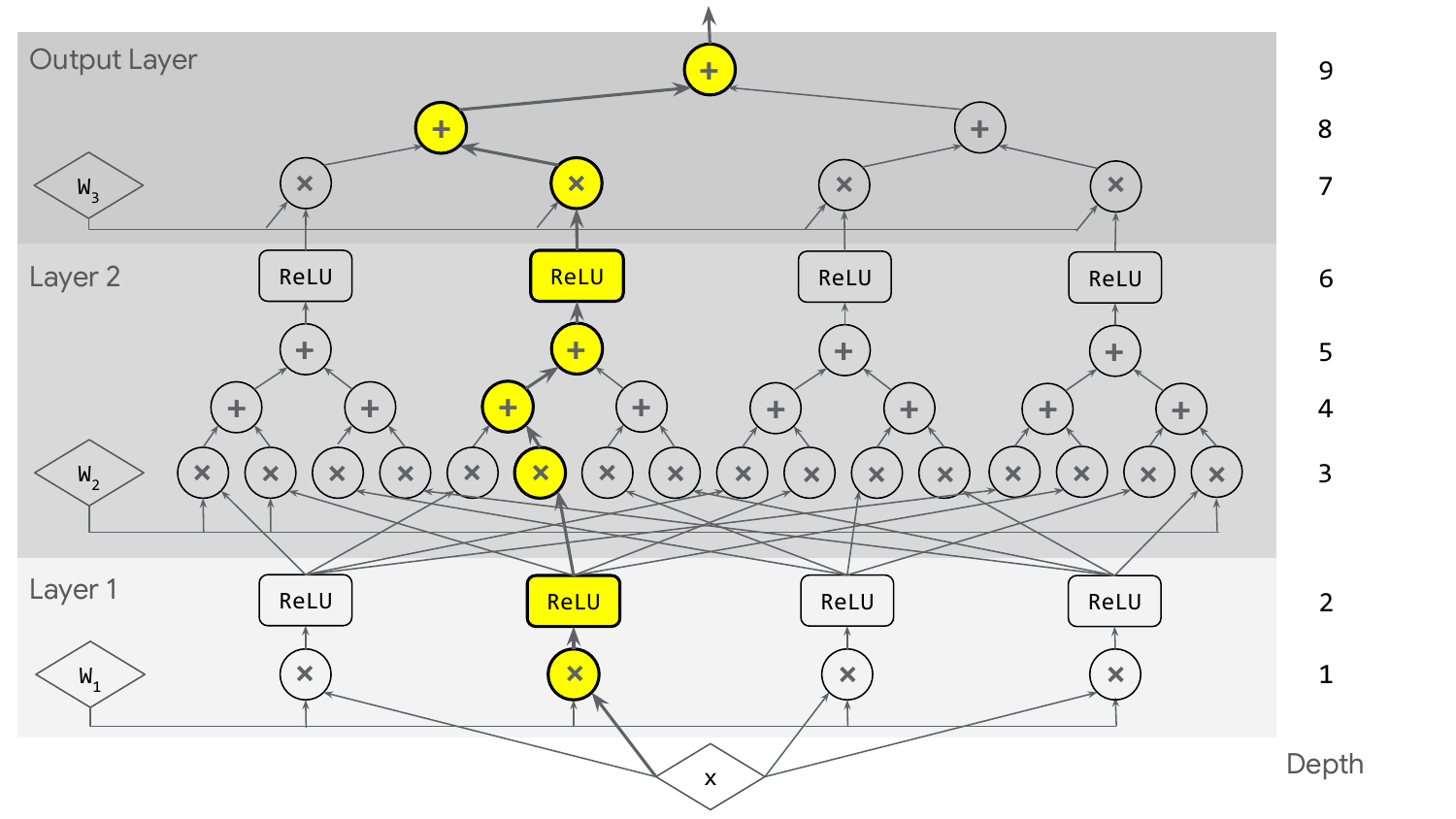}
  \caption{Serial depth of a circuit computing an MLP with two hidden layers, given a single input and a single output. The highlighted path is the longest in the circuit, and has length 9, determining the serial depth. The paths from the weights $W_2$ and $W_3$ to the output have a lower length, and so do not contribute to the serial depth.} 
  \label{fig:two-layer-mlp}
\end{figure}

To understand this depth calculation method, we now manually compute the depth of a simple two hidden-layer MLP with a single input dimension, a single output dimension, and two hidden layers with dimension 4.
The activations in the hidden layers will be ReLUs.

For this example the serial depth can be computed layer by layer, as illustrated in \cref{fig:two-layer-mlp}:

\begin{itemize}
\item The first hidden layer has 4 neurons, and each neuron computes a product with
  the input. These can happen in parallel which gives depth 1. This is followed
  by a ReLU activation, so the total depth contribution is $1 + 1 = 2$.
\item The second hidden layer has 4 neurons, and computes a product with each input
  (depth 1), a sum over 4 inputs (depth 2), and a ReLU activation (depth 1).
  The total depth contribution is $1 + \text{ceil}(\log(4)) + 1 = 4$.
\item The output layer has 1 output, and computes a product for each input and then
  a sum over 4 inputs. The depth contribution is $1 + \text{ceil}(\log(4)) = 3$.
\end{itemize}
Hence, an upper bound on the serial depth of the MLP is the sum of these contributions: $2 + 4 + 3 = 9$.

\subsection{Depth Calculations for Gemma 3} \label{sec:gemma-3-depth-by-hand}

For a significantly more complicated example, we calculate upper bounds on the opaque serial depth by hand for the models from the Gemma 3 family~\citep{team2025gemma}, treating the input and output tokens as the interpretable nodes. We provide the full calculation in \cref{appendix:hand-examples} and summarize the results in \cref{tab:gemma-depth-summary}.

\begin{table}[ht]
\centering
\begin{tabular}{|l|l|l|}
\hline
\textbf{Model} & \textbf{Final Depth Formula} & \textbf{Total Depth (T=T\_max)} \\
\hline
\textbf{Gemma 3 1B} & \textbf{$4370 + 8\cdot\log_2 T$} & \textbf{4,490} \\
\textbf{Gemma 3 4B} & \textbf{$6036 + 10\cdot\log_2 T$} & \textbf{6,206} \\
\textbf{Gemma 3 12B} & \textbf{$8482 + 16\cdot\log_2 T$} & \textbf{8,754} \\
\textbf{Gemma 3 27B} & \textbf{$11322 + 20\cdot\log_2 T$} & \textbf{11,662} \\
\hline
\end{tabular}
\caption{Upper bounds on the serial depth of Gemma 3 models at maximum sequence length.}
\label{tab:gemma-depth-summary}
\end{table}

\subsection{Asymptotic bounds on depths for different architectures}
\begin{figure}[t]
    \centering
    \begin{subfigure}[t]{0.48\textwidth}
        \centering
        \includegraphics[trim=0 12 180 0, clip, width=\textwidth]{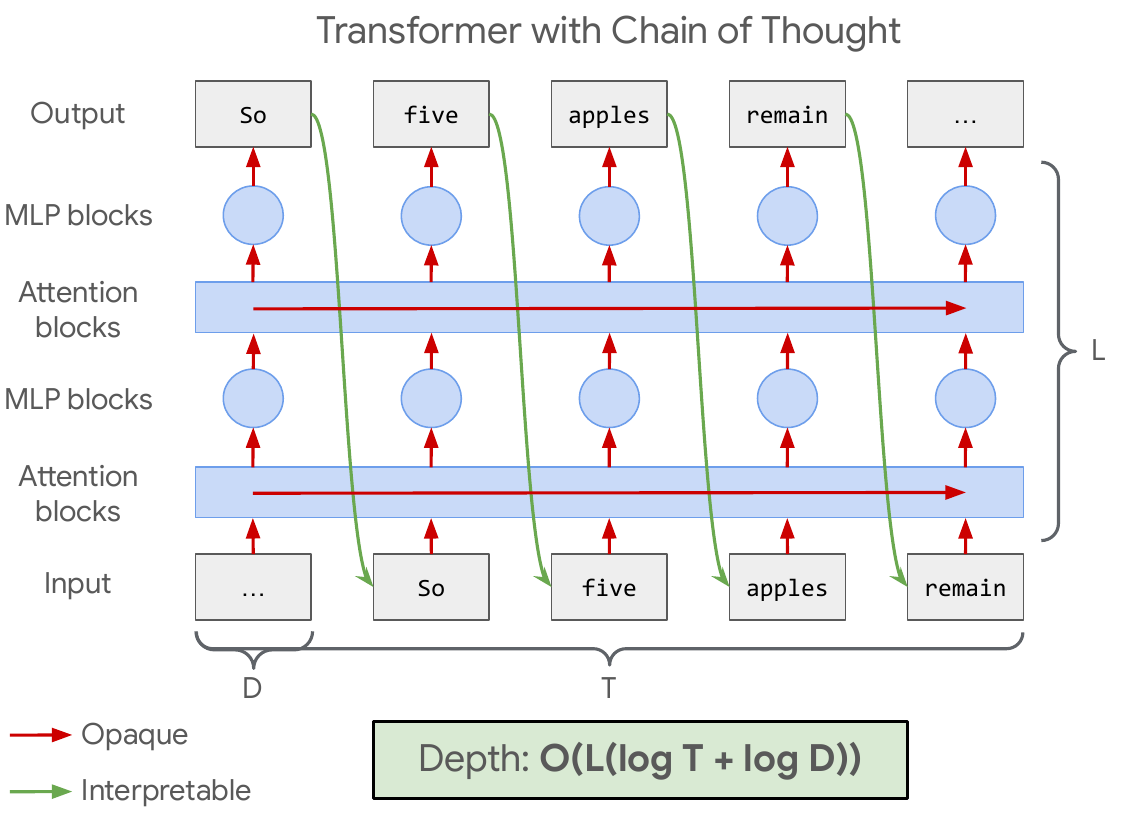}
        \caption{In a standard Transformer, opaque paths can only go ``up'' or ``right'', with linear dependence on ``up'' steps and logarithmic dependence on ``right'' steps.}
        \label{subfig:transformer}
    \end{subfigure}%
    \hfill
    \begin{subfigure}[t]{0.48\textwidth}
        \centering
        \includegraphics[trim=0 12 180 0, clip, width=\textwidth]{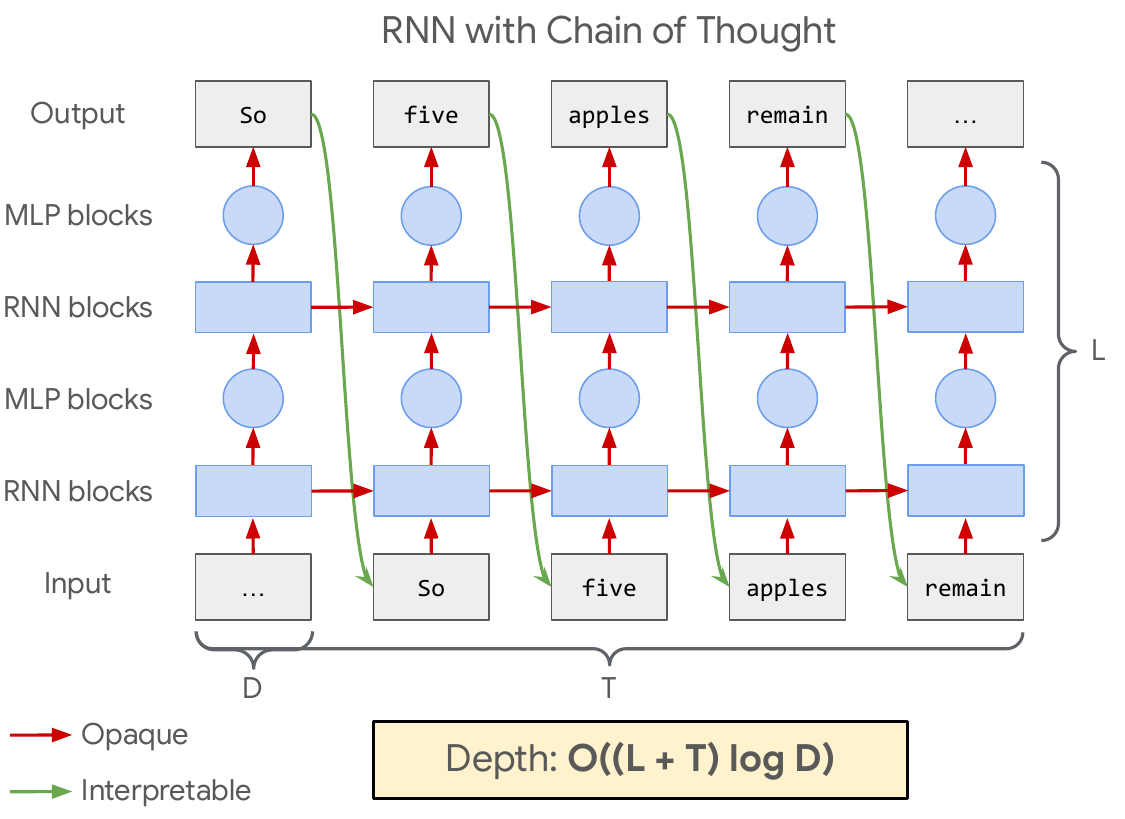}
        \caption{With RNN blocks, opaque paths can still only go ``up'' or ``right''. However, there is now a linear dependence on ``right'' steps.}
        \label{subfig:rnn}
    \end{subfigure}

    \vspace{12pt}
    
    \begin{subfigure}[t]{0.48\textwidth}
        \centering
        \includegraphics[trim=0 12 180 0, clip, width=\textwidth]{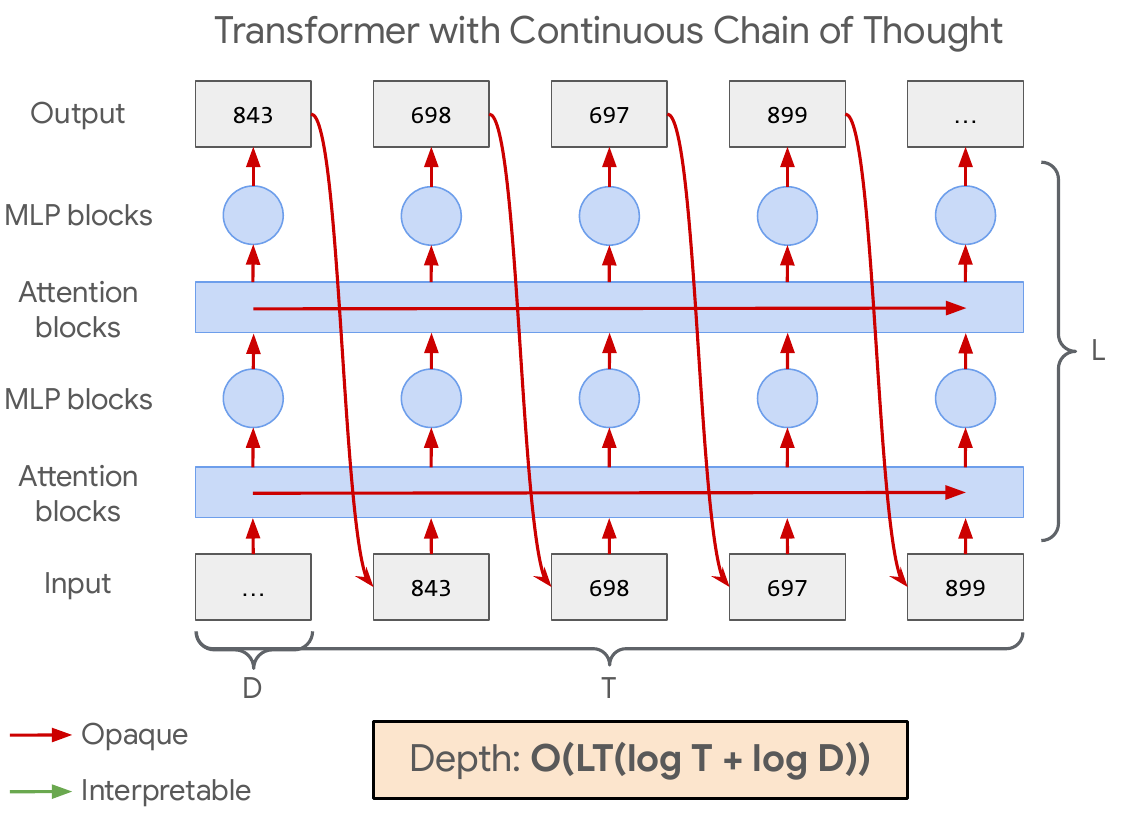}
        \caption{Once the chain of thought is not interpretable, opaque paths can also go ``down'' in addition to ``up'' or ``right'', drastically increasing opaque serial depth.}
        \label{subfig:continuous-cot}
    \end{subfigure}%
    \hfill
    \begin{subfigure}[t]{0.48\textwidth}
        \centering
        \includegraphics[trim=0 12 180 0, clip, width=\textwidth]{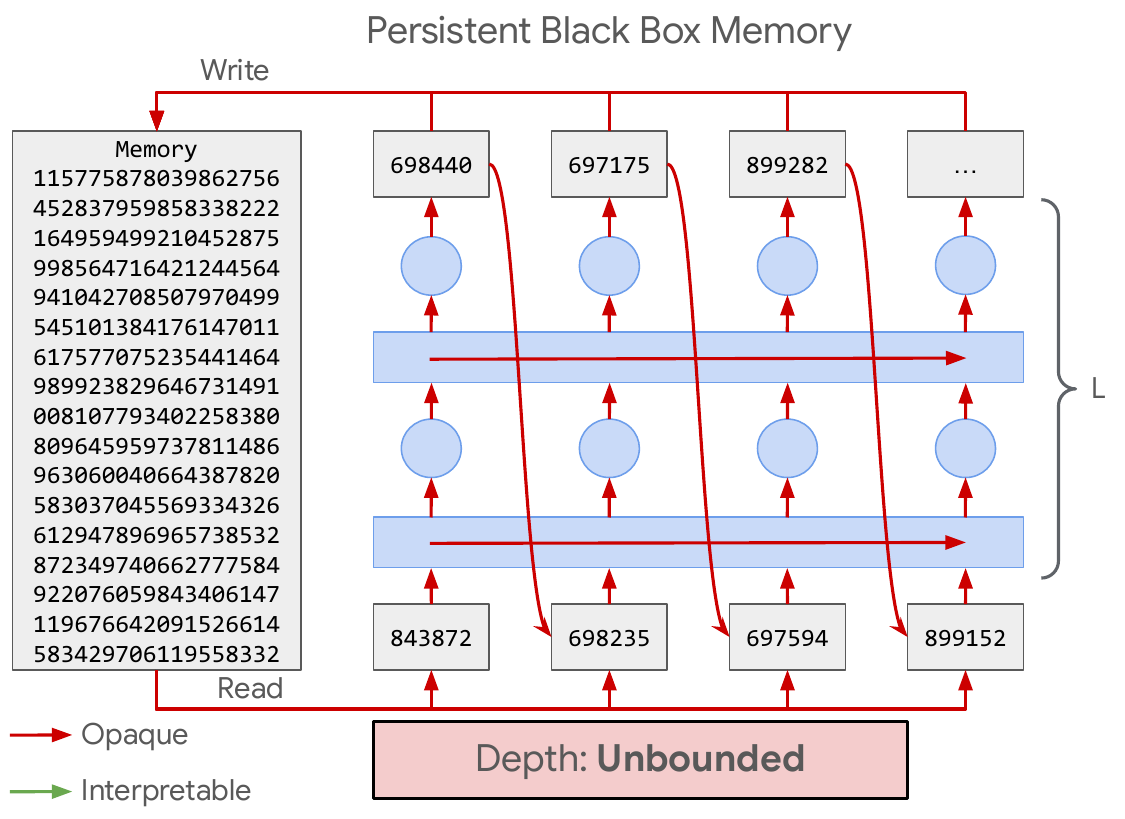}
        \caption{Contents in persistent black box memory could have been produced with arbitrary opaque serial computation, making the depth unbounded.}
        \label{subfig:memory}
    \end{subfigure}
    \caption{Asymptotic analysis of opaque serial depth for various architectures shows that architectural choices can make a large difference to the amount of opaque serial computation that can be done.}
    \label{fig:asymptotics}
\end{figure}

With a definition of opaque serial depth in hand, we can identify asymptotic bounds for a given architecture, allowing us to predict how opaque serial depth will change as models are scaled up.
For the architectures we discuss here, we let $D$ be the dimension of the activations, $L$ be the number of layers, and $T$ be the number of tokens. The $L$ layers count all types of layers, such as attention, MLPs, mixture of experts, and RNNs. The $T$ tokens can be decomposed into $T_{in}$ input tokens, $T_{cot}$ chain of thought tokens (where applicable), and $T_{out}$ output tokens. Input and output tokens are always considered interpretable. We illustrate many of these architectures and their depths in \cref{fig:asymptotics}.

\paragraph{Autoregressive Transformers (\cref{subfig:transformer}).} In an autoregressive Transformer, every token is an interpretable node, and so an opaque path is limited to the forward pass required to produce a single token. Since each output token $i$ is computed by a forward pass on an input sequence of length $i-1$, the longest opaque path will be for the final output token, when using the final sequence length $T$.

Each MLP block (or mixture of experts block) operates on a single token, computing a constant number of simple operations over activation vectors of size $D$, which requires depth $O(\log D)$. The attention operation also includes such operations of depth $O(\log D)$, as well as a sum over $T$ numbers when calculating the attention pattern, which requires depth $\log_2 T$.

Thus, every layer has depth $O(\log T + \log D)$. Since the overall forward pass is a composition of $L$ layers, the overall opaque serial depth is $O(L(\log T + \log D))$.

\paragraph{Recurrent Networks (\cref{subfig:rnn}).} Instead of using attention layers to pass information from earlier tokens to later tokens, one could instead use RNN blocks. Each application involves a constant number of simple operations over activation vectors of size $D$ and so requires depth $O(\log D)$.

Any overall opaque path through the model then consists of two types of steps: processing the next deeper layer (going up in \cref{subfig:rnn}), or moving to the next token via an RNN block (going right in \cref{subfig:rnn}). There can be at most $L$ steps of the first type, and $T$ steps of the second type, and each step has depth $O(\log D)$. Thus, the overall opaque serial depth is $O((L + T)\log D)$.

\paragraph{Continuous Latent Chain of Thought (\cref{subfig:continuous-cot}).}
There have been proposals for continuous latent chain of thought \citep{hao2024training}. Rather than sampling tokens from a language model $f_{\theta}$, one instead uses the final hidden state as a continuous representation of the current ``thought'' and feeds this continuous input directly back into the network as the input embedding, bypassing tokens entirely.

It is unclear whether the new continuous thought tokens should be considered interpretable: for example, perhaps they can still be understood as discrete tokens via the logit lens~\citep{nostalgebraist2020interpreting}. For the purposes of this section, we will assume that they are \textbf{not} interpretable. Thus, only the input and output tokens are interpretable.

In this case, the overall computation is the composition of the backing neural network $f_{\theta}$ for $T_{cot}$ steps. Assuming that we are using a Transformer architecture, each invocation of the $f_{\theta}$ has depth $O(L(\log T + \log D))$. Thus, the overall depth is $O(L T_{cot} (\log T + \log D))$. This illustrates how our definition of opaque serial depth naturally assigns larger depth to longer computations that do not produce any interpretable intermediate steps.

\paragraph{Persistent Black Box Memory (\cref{subfig:memory}).} A language model could be equipped with memory that persists across individual queries. If this memory is interpretable, this does not change the opaque serial depth. However, if it is uninterpretable or ``black box'', for example because it stores activation vectors, this can significantly increase opaque serial depth.

In the case where the language model can \emph{read} from the memory, carry out further opaque processing, and then \emph{write} back to the memory, the amount of opaque serial computation put into any individual item stored in memory can grow continually. Thus, the opaque serial depth cannot be bounded in terms of $L$, $T$ and $D$. It could be bounded by including other quantities such as the total number of invocations of the language model or the total number of writes to the memory system, though such bounds would likely be too high to provide any kind of assurance about our understanding of the computation.

\paragraph{Text Diffusion Models.}
For a text diffusion model \citep{austin2021structured,li2022diffusion,nie2025large}, a single neural network $f_{\theta}$ is used to produce $T$ tokens by an iterative diffusion process. The process proceeds in a series of diffusion steps, where in step $t$ a sequence $s_t$ of $T$ tokens are produced in parallel, and then the tokens $s_t$ are fed as inputs to $f_{\theta}$ in step $t+1$.
If we view any tokens output as interpretable, then a $k$-step text-diffusion process producing $T$ tokens has depth $\text{Depth}(f_{\theta})$, because each interpretable node for a token produced at step $i$ in the circuit simply applies $f_{\theta}$ to the interpretable nodes corresponding to the output tokens of the previous step.

\section{Automating Serial Depth Calculations} \label{sec:automated-calculation}

While we give in-depth examples of hand-calculations in \cref{appendix:hand-examples}, they can become quite tedious. To automate this process we also implemented a depth upper bound calculator in JAX~\citep{jax2018github}.

\subsection{Approach}

We apply Algorithm~\ref{alg:opaque-serial-depth} to the jaxpr intermediate language representation of the neural network. In particular, for each jaxpr that occurs in a neural network implemented in JAX, we first determine by hand the depth of the corresponding mathematical operation, and hard-code this into our implementation. The algorithm then recursively traverses the tree of jaxprs to compute the maximum depth along any path.

Overall, the number of jaxprs that need to be considered is not too large, with approximately 75 being sufficient to run a depth calculation for the full Gemma 3 architecture. The vast majority of these are quite simple depth 0 operations, such as reshaping or other indexing manipulations which correspond to the wiring of the induced circuit, or depth 1 operations, typically consisting of elementary coordinate-wise mathematical operations. We implement 7 further jaxprs that correspond to associative binary operations over multiple inputs, and 11 more that require special handling due to recursive calling of further jaxprs and other special-case operations. We open-source our code at \url{https://github.com/google-deepmind/serial_depth}.

\subsection{Comparison to manual calculations}

Our automated method calculates the opaque serial depth of the circuit implied by the jaxpr representation of the neural network. However, the jaxpr representation is not optimized to minimize serial depth. As a result, it often misses optimizations to the circuit that can tighten the upper bound on the opaque serial depth of the neural network.

\begin{figure}[t]
    \centering
    \begin{subfigure}[t]{\textwidth}
        \centering
        \includegraphics[trim=0 0 72 156, clip, width=\textwidth]{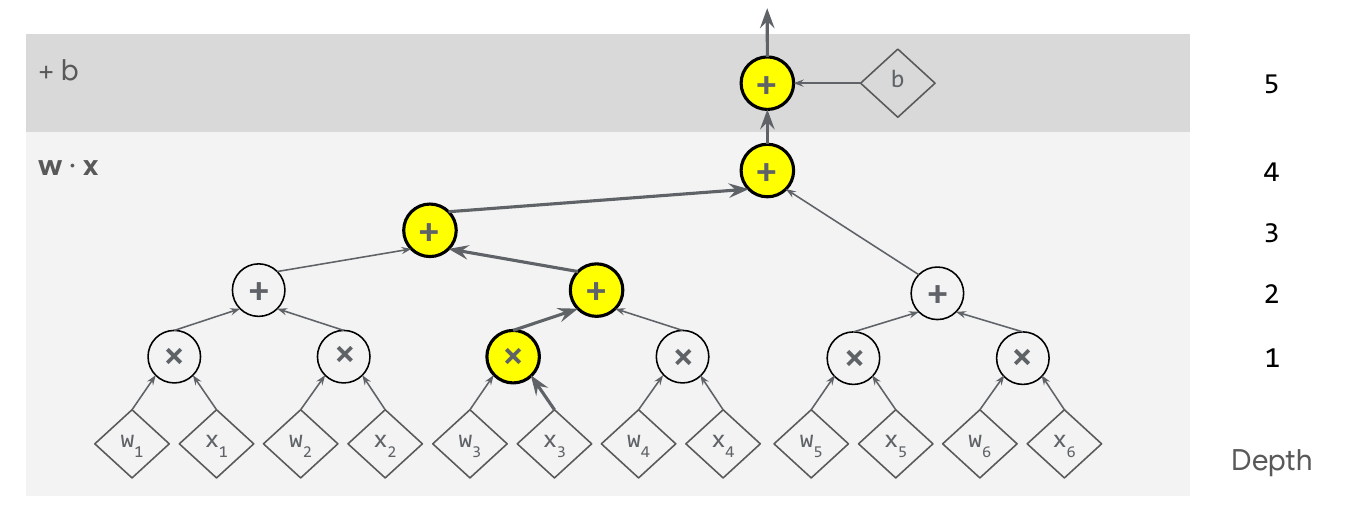}
        \caption{A tree traversal may compute a depth of 4 for $\bm{w}\cdot\bm{x}$, and add a depth of 1 for $+ \; b$.}
    \end{subfigure}
    \begin{subfigure}[t]{\textwidth}
        \centering
        \includegraphics[trim=0 0 72 204, clip, width=\textwidth]{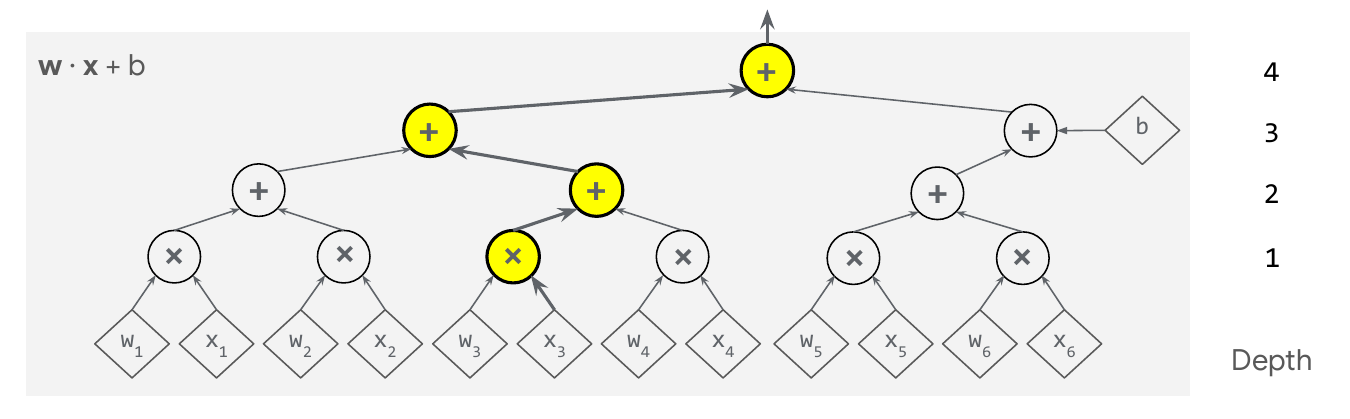}
        \caption{Folding in the addition of the bias $b$ into the dot product sum reduces the depth by 1.}
    \end{subfigure}
    \caption{Manual optimizations can produce tighter upper bounds on the serial depth of a neural network. When computing $\bm{w} \cdot \bm{x} + b$ where the vectors have dimension 6, an automated calculation can produce a depth of 5, but folding in the addition of the bias into the dot product sum produces a shallower circuit of depth 4.}
    \label{fig:optimizing-wx-b}
\end{figure}

One common missed optimization is illustrated in \cref{fig:optimizing-wx-b}. A common pattern in neural networks is a matrix multiplication or dot product, followed by the addition of a bias. If the intermediate representation treats the addition of the bias as a separate step, this will typically increase the calculated serial depth. However, usually the addition of the bias can be folded into the sums involved in the matrix multiplication or dot product, thus avoiding the additional depth. This is analogous to the common trick of adding a feature that is always set to 1 to the data, such that the weight corresponding to that feature functions equivalently to a bias.

To quantify the impact of these missed optimizations, we run our depth calculator on Gemma 3 models, and compare the results to the depths calculated by hand in \cref{sec:gemma-3-depth-by-hand}. We provide the results in \cref{tab:gemma-depth-comparison}. For both cases, we report the depth at the maximum sequence length supported by each model.

\begin{table}[ht]
\centering
\begin{tabular}{|l|l|l|l|}
\hline
\textbf{Model} & \textbf{By-Hand Depth} & \textbf{JAX Calculator Depth} & \textbf{JAX Overestimation}\\
\hline
\textbf{Gemma 3 1B} & 4,490 & 5,728 & $1.28\times$\\
\textbf{Gemma 3 4B} & 6,206 & 7,958 & $1.28\times$\\
\textbf{Gemma 3 12B} & 8,754 & 11,268 & $1.29\times$ \\
\textbf{Gemma 3 27B} & 11,662 & 14,856 & $1.27\times$ \\
\hline
\end{tabular}
\caption{Hand-calculated upper bounds on the serial depth of Gemma 3 models versus JAX depth calculator upper bounds at maximum sequence length.}
\label{tab:gemma-depth-comparison}
\end{table}

As expected, the JAX depth calculator overestimates the serial depth relative to the by-hand calculations, by around $28\%$ for most models in the family. We view a $28\%$ overhead as an excellent trade-off for the efficiency of the JAX depth calculator. The by-hand calculations can taken several hours of human effort to carefully verify for a new architecture, while the calculator runs in a few seconds. The by-hand calculations are also more error prone, and could miss some architectural details, whereas the JAX depth calculator works directly on an actual implementation of the neural network. Of course, if one is interested in the tightest possible upper bounds on the depth of a neural network, the by-hand calculations will be more effective.

\begin{figure}[th] 
  \centering %
  \includegraphics[width=0.8\textwidth]{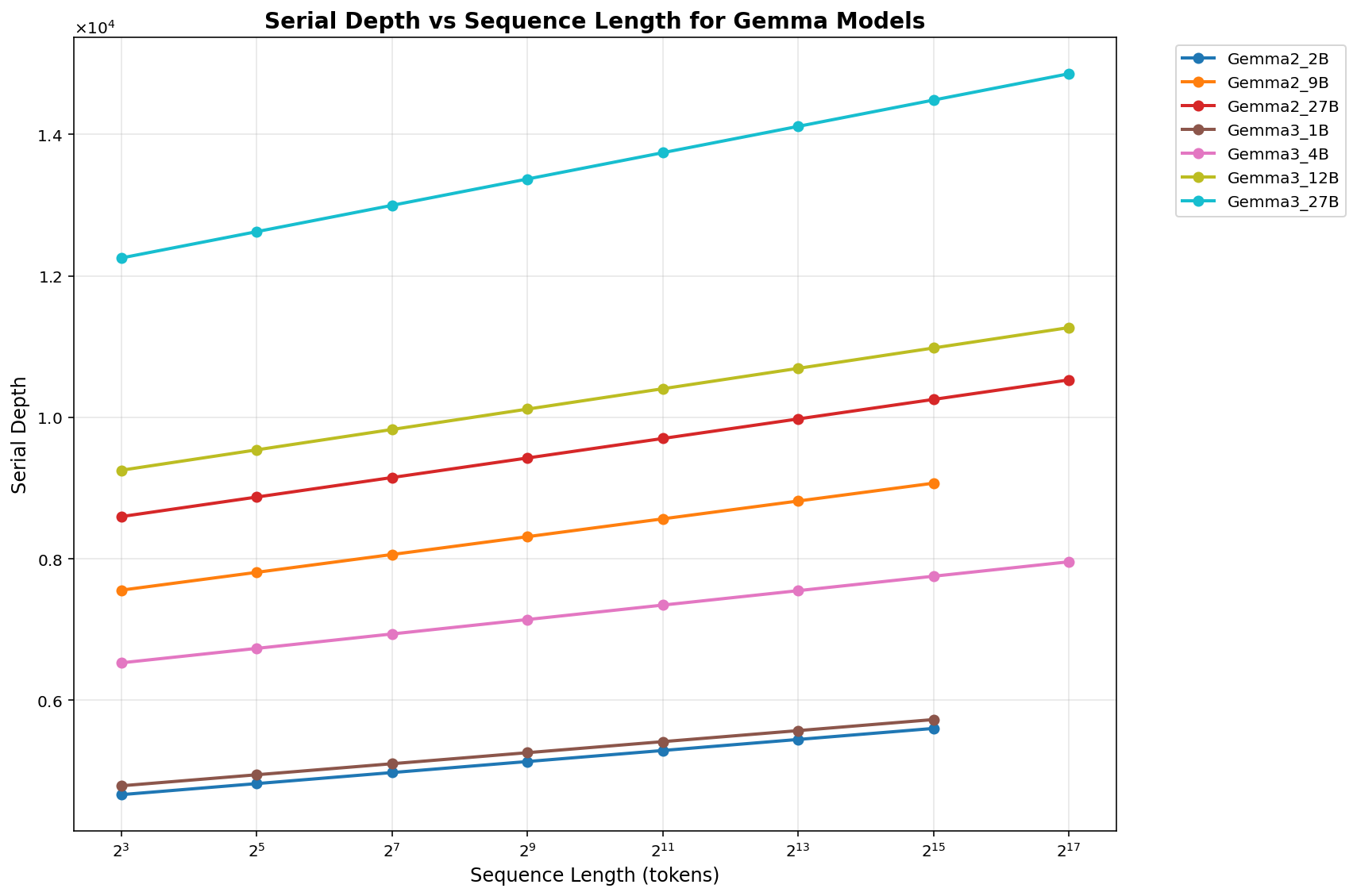}
  \caption{Automatic depth calculations for Gemma models across different sequence lengths.} 
  \label{fig:gemma-depth}
\end{figure}

The by-hand calculations summarized in \cref{tab:gemma-depth-summary} predict logarithmic scaling of serial depth with sequence length for Gemma 3 models. In \cref{fig:gemma-depth}, we plot the JAX-calculated depth as a function of sequence length for various models in the Gemma 2 and 3 families, confirming this predicted logarithmic scaling.

\subsection{Application to Mixture-of-Experts}

Since the automated calculations are so cheap, and only require a model architecture (rather than a fully trained model), it is easy to get quick qualitative evidence about the impact of various architectural changes on the opaque serial depth.

As a proof of concept, we implement a vanilla Mixture-of-Experts (MoE) Transformer model in JAX with 11B active parameters and 91B total parameters (full details in \cref{tab:moe-architechture}). The JAX calculator produces a depth of 4,096 for this model, which is significantly lower than the JAX-calculated depth of 11,268 for Gemma 3 12B, and even the by-hand calculation of 8,754 (\cref{tab:gemma-depth-comparison}), suggesting that Mixture-of-Experts likely decreases serial depth relative to dense architectures.

\begin{table}[ht]
\centering
\begin{tabular}{|l|l|}
\hline
\textbf{Parameter} & \textbf{Value} \\
\hline
  Vocabulary size & 151936 \\
  Hidden dim & 2048 \\
  Num heads & 16 \\
  Num layers & 28 \\
  Experts per layer & 64 \\
  Experts per token & 8 \\
  Sequence length & 512 \\
  Total parameters & 91.32B \\
  Active parameters & 11.42B \\
\hline
\textbf{Total Calculated Depth} & \textbf{4096} \\
\hline
\end{tabular}
\caption{Parameters of the MoE Transformer architecture and total depth output by the JAX depth calculator.}
\label{tab:moe-architechture}
\end{table}

\section{Limitations} \label{sec:limitations}

\paragraph{Interpretable ``nodes'' only capture one aspect of transparency.} When considering a computation expressed as a circuit, an individual node in that circuit does not represent the entire algorithm, but rather a snapshot of part of the computational state~\citep{levy2025state}. In principle, this can give us some insight into the computation, but would be far from an explanation of the algorithm involved.

However, since LLMs often imitate and improve upon written human reasoning, the chain of thought often lays out the reasoning explicitly, and so \emph{can} serve as an explanation. This is not captured by our formalization of opaque serial depth, and thus other architectures (e.g. text diffusion models) may lose this property despite having similar opaque serial depth.

\paragraph{Definition of interpretability.} A key limitation of the method is the need to specify which nodes are ``interpretable''. While we discuss some ideas for operationalizing interpretability in \cref{sec:interpretable-nodes} and \cref{appendix:interpretable-nodes}, ultimately there is not a clean, formal definition that captures what we want, and these determinations will likely involve significant judgment calls.

\paragraph{Quantification over circuits.} Depth is defined via a minimum over polynomially sized circuits (\cref{sec:serial-depth-nn}). While this captures the benefits of serial reasoning in theory (\cref{sec:why-circuit-depth}), theory may not apply perfectly to practice, which is often highly inefficient relative to the theoretical optimum.

For example, any function can be implemented by an input-output lookup table of exponential size and linear depth. Such a circuit could have a lower depth than the standard implementation for e.g. continuous chain of thought (\cref{subfig:continuous-cot}). However, it is prima facie unreasonable to say that simply because we could cache the full input-output behavior of such an architecture, that means it is not doing sophisticated serial reasoning.

While the polynomial size restriction rules out lookup tables, there may be other such cases we have not thought of. If opaque serial depth is used for decision making with practical systems, it may be prudent to only quantify over ``natural'' or ``reasonable'' circuits (determined by expert judgment).

\paragraph{New operations for automated depth calculations.} The automated depth calculator (\cref{sec:automated-calculation}) only supports a subset of JAX operations. For models that use unsupported operations, users will have to extend the calculator with a new formula for that operation produced by hand.

\section{Conclusion} \label{sec:conclusion}

We would like to build AI systems that perform sophisticated \emph{human-understandable} reasoning. We capture the potential for sophisticated reasoning using the concept of serial (circuit) depth. With this language, we want AI systems with high serial depth, but low \emph{opaque} serial depth.

We provide an algorithm that calculates upper bounds on the opaque serial depth of a neural network, enabling both theoretical analysis by hand of proposed architectures, and an automated method for concretely instantiated models. This provides a standardized way to get quick concrete estimates that inform our understanding of the extent to which a proposed architecture enables sophisticated human-understandable reasoning.

We hope that opaque serial depth will aid in the development of increasingly capable AI architectures while retaining a similar level of transparency as we have with chain of thought currently.

\bibliography{depth}

@article{papadimitriou1987complexity,
  title={The complexity of Markov decision processes},
  author={Papadimitriou, Christos H and Tsitsiklis, John N},
  journal={Mathematics of operations research},
  volume={12},
  number={3},
  pages={441--450},
  year={1987},
  publisher={INFORMS}
}

@article{hahn2020theoretical,
  title={Theoretical limitations of self-attention in neural sequence models},
  author={Hahn, Michael},
  journal={Transactions of the Association for Computational Linguistics},
  volume={8},
  pages={156--171},
  year={2020},
  publisher={MIT Press One Rogers Street, Cambridge, MA 02142-1209, USA journals-info~…}
}

@article{merrill2022saturated,
  title={Saturated transformers are constant-depth threshold circuits},
  author={Merrill, William and Sabharwal, Ashish and Smith, Noah A},
  journal={Transactions of the Association for Computational Linguistics},
  volume={10},
  pages={843--856},
  year={2022},
  publisher={MIT Press One Broadway, 12th Floor, Cambridge, Massachusetts 02142, USA~…}
}

@inproceedings{liu2024Chain,
  author       = {Zhiyuan Liu and
                  Hong Liu and
                  Denny Zhou and
                  Tengyu Ma},
  title        = {Chain of Thought Empowers Transformers to Solve Inherently Serial
                  Problems},
  booktitle    = {The Twelfth International Conference on Learning Representations,
                  {ICLR} 2024, Vienna, Austria, May 7-11, 2024},
  publisher    = {OpenReview.net},
  year         = {2024},
  url          = {https://openreview.net/forum?id=3EWTEy9MTM},
}

@misc{jax2018github,
  author = {James Bradbury and Roy Frostig and Peter Hawkins and Matthew James Johnson and Chris Leary and Dougal Maclaurin and George Necula and Adam Paszke and Jake Vander{P}las and Skye Wanderman-{M}ilne and Qiao Zhang},
  title = {{JAX}: composable transformations of {P}ython+{N}um{P}y programs},
  url = {http://github.com/jax-ml/jax},
  version = {0.3.13},
  year = {2018},
}

@article{team2025gemma,
  title={Gemma 3 technical report},
  author={Team, Gemma and Kamath, Aishwarya and Ferret, Johan and Pathak, Shreya and Vieillard, Nino and Merhej, Ramona and Perrin, Sarah and Matejovicova, Tatiana and Ram{\'e}, Alexandre and Rivi{\`e}re, Morgane and others},
  journal={arXiv preprint arXiv:2503.19786},
  year={2025}
}

@inproceedings{Beame1984LogDC,
  title={Log Depth Circuits for Division and Related Problems},
  author={Paul Beame and Stephen A. Cook and H. James Hoover},
  booktitle={SIAM journal on computing (Print)},
  year={1984},
  url={https://api.semanticscholar.org/CorpusID:535657}
}

@article{austin2021structured,
  title={Structured denoising diffusion models in discrete state-spaces},
  author={Austin, Jacob and Johnson, Daniel D and Ho, Jonathan and Tarlow, Daniel and Van Den Berg, Rianne},
  journal={Advances in neural information processing systems},
  volume={34},
  pages={17981--17993},
  year={2021}
}

@article{li2022diffusion,
  title={Diffusion-lm improves controllable text generation},
  author={Li, Xiang and Thickstun, John and Gulrajani, Ishaan and Liang, Percy S and Hashimoto, Tatsunori B},
  journal={Advances in neural information processing systems},
  volume={35},
  pages={4328--4343},
  year={2022}
}

@article{hao2024training,
  title={Training large language models to reason in a continuous latent space},
  author={Hao, Shibo and Sukhbaatar, Sainbayar and Su, DiJia and Li, Xian and Hu, Zhiting and Weston, Jason and Tian, Yuandong},
  journal={arXiv preprint arXiv:2412.06769},
  year={2024}
}

@article{korbak2025chain,
  title={Chain of thought monitorability: A new and fragile opportunity for ai safety},
  author={Korbak, Tomek and Balesni, Mikita and Barnes, Elizabeth and Bengio, Yoshua and Benton, Joe and Bloom, Joseph and Chen, Mark and Cooney, Alan and Dafoe, Allan and Dragan, Anca and others},
  journal={arXiv preprint arXiv:2507.11473},
  year={2025}
}

@article{emmons2025chain,
  title={When chain of thought is necessary, language models struggle to evade monitors},
  author={Emmons, Scott and Jenner, Erik and Elson, David K and Saurous, Rif A and Rajamanoharan, Senthooran and Chen, Heng and Shafkat, Irhum and Shah, Rohin},
  journal={arXiv preprint arXiv:2507.05246},
  year={2025}
}

@article{zhang2025soft,
  title={Soft thinking: Unlocking the reasoning potential of llms in continuous concept space},
  author={Zhang, Zhen and He, Xuehai and Yan, Weixiang and Shen, Ao and Zhao, Chenyang and Wang, Shuohang and Shen, Yelong and Wang, Xin Eric},
  journal={arXiv preprint arXiv:2505.15778},
  year={2025}
}

@article{zhuang2025text,
  title={Text generation beyond discrete token sampling},
  author={Zhuang, Yufan and Liu, Liyuan and Singh, Chandan and Shang, Jingbo and Gao, Jianfeng},
  journal={arXiv preprint arXiv:2505.14827},
  year={2025}
}

@article{wu2025llms,
  title={Llms are single-threaded reasoners: Demystifying the working mechanism of soft thinking},
  author={Wu, Junhong and Lu, Jinliang and Ren, Zixuan and Hu, Gangqiang and Wu, Zhi and Dai, Dai and Wu, Hua},
  journal={arXiv preprint arXiv:2508.03440},
  year={2025}
}

@article{ziegler2019fine,
  title={Fine-tuning language models from human preferences},
  author={Ziegler, Daniel M and Stiennon, Nisan and Wu, Jeffrey and Brown, Tom B and Radford, Alec and Amodei, Dario and Christiano, Paul and Irving, Geoffrey},
  journal={arXiv preprint arXiv:1909.08593},
  year={2019}
}

@article{ouyang2022training,
  title={Training language models to follow instructions with human feedback},
  author={Ouyang, Long and Wu, Jeffrey and Jiang, Xu and Almeida, Diogo and Wainwright, Carroll and Mishkin, Pamela and Zhang, Chong and Agarwal, Sandhini and Slama, Katarina and Ray, Alex and others},
  journal={Advances in neural information processing systems},
  volume={35},
  pages={27730--27744},
  year={2022}
}

@article{nie2025large,
  title={Large language diffusion models},
  author={Nie, Shen and Zhu, Fengqi and You, Zebin and Zhang, Xiaolu and Ou, Jingyang and Hu, Jun and Zhou, Jun and Lin, Yankai and Wen, Ji-Rong and Li, Chongxuan},
  journal={arXiv preprint arXiv:2502.09992},
  year={2025}
}

@article{jaech2024openai,
  title={Openai o1 system card},
  author={Jaech, Aaron and Kalai, Adam and Lerer, Adam and Richardson, Adam and El-Kishky, Ahmed and Low, Aiden and Helyar, Alec and Madry, Aleksander and Beutel, Alex and Carney, Alex and others},
  journal={arXiv preprint arXiv:2412.16720},
  year={2024}
}

@article{guan2025monitoring,
  title={Monitoring Monitorability},
  author={Guan, Melody and Wang, Miles and Carroll, Micah and Dou, Zehao and Wei, Annie and Williams, Marcus and Arnav, Benjamin and Huizinga, Joost and Kivlichan, Ian and Glaese, Mia and Pachocki, Jakub and Baker, Bowen},
  journal={arXiv preprint arXiv:2512.18311},
  year={2025}
}

@article{levy2025state,
  title={State over Tokens: Characterizing the Role of Reasoning Tokens},
  author={Levy, Mosh and Elyoseph, Zohar and Ravfogel, Shauli and Goldberg, Yoav},
  journal={arXiv preprint arXiv:2512.12777},
  year={2025}
}

@misc{roger2025reasoning,
  title={Do Reasoning Models Use Their Scratchpad like We Do? {E}vidence from Distilling Paraphrases},
  author={Roger, Fabien},
  year={2025},
  url={https://alignment.anthropic.com/2025/distill-paraphrases/}
}

@article{emmons2025pragmatic,
  title={A Pragmatic Way to Measure Chain-of-Thought Monitorability},
  author={Emmons, Scott and Zimmermann, Roland S and Elson, David K and Shah, Rohin},
  journal={arXiv preprint arXiv:2510.23966},
  year={2025}
}

@article{farquhar2025mona,
  title={Mona: Myopic optimization with non-myopic approval can mitigate multi-step reward hacking},
  author={Farquhar, Sebastian and Varma, Vikrant and Lindner, David and Elson, David and Biddulph, Caleb and Goodfellow, Ian and Shah, Rohin},
  journal={arXiv preprint arXiv:2501.13011},
  year={2025}
}

@misc{nostalgebraist2020interpreting,
  title={interpreting {GPT}: the logit lens},
  author={nostalgebraist},
  year={2020},
  url={https://www.lesswrong.com/posts/AcKRB8wDpdaN6v6ru/interpreting-gpt-the-logit-lens}
}

@misc{deng2025cot,
  title={{CoT} May Be Highly Informative Despite “Unfaithfulness”},
  author={Deng, Amy and Von Arx, Sydney and Snodin, Ben and Kunnavakkam, Sudarsh and Lanham, Tamera},
  year={2025},
  url={https://metr.org/blog/2025-08-08-cot-may-be-highly-informative-despite-unfaithfulness/}
}

@article{vaswani2017attention,
  title={Attention is all you need},
  author={Vaswani, Ashish and Shazeer, Noam and Parmar, Niki and Uszkoreit, Jakob and Jones, Llion and Gomez, Aidan N and Kaiser, {\L}ukasz and Polosukhin, Illia},
  journal={Advances in neural information processing systems},
  volume={30},
  year={2017}
}

@article{zamir2024undetectable,
  title={Undetectable steganography for language models},
  author={Zamir, Or},
  journal={Transactions on Machine Learning Research},
  year={2024}
}

@inproceedings{christ2024undetectable,
  title={Undetectable watermarks for language models},
  author={Christ, Miranda and Gunn, Sam and Zamir, Or},
  booktitle={The Thirty Seventh Annual Conference on Learning Theory},
  pages={1125--1139},
  year={2024},
  organization={PMLR}
}

\newpage
\appendix

\section{Determining whether a node is interpretable} \label{appendix:interpretable-nodes}

As discussed in \cref{sec:interpretable-nodes}, opaque serial depth calculations are very sensitive to the choice of which subset of nodes are considered interpretable. In this section, while we do not provide a precise definition of interpretable nodes, we discuss some strategies for deciding whether a node should be considered interpretable.

For a human to interpret an intermediate node, there must be some way to convert the node into a form that humans can understand. For simplicity, we will only consider translations to natural language text. Often this transformation will just be the identity function (e.g. an LLM's chain of thought is already in natural language) but this is not always the case (e.g. residual stream activations might be transformed to text via the logit lens~\citep{nostalgebraist2020interpreting}).

One very straightforward approach is to simply check that the resulting text looks like coherent language, and covers the reasoning that we expect~\citep{emmons2025pragmatic}. However, this is just a sanity check, and would not be suitable to certify that a node is interpretable.
Instead, we consider two properties that could much more strongly suggest that the node overall is interpretable. The first property is the extent to which the nodes enable answering questions about the overall computation, and the second property is the extent to which the information contained in the node is understandable to humans.

\subsection{Question answering}

Intuitively, the point of interpretability is to allow humans to understand what the model is doing. One way to operationalize ``understanding'' a node is to say that we should be able to answer arbitrary questions about the node. Thus, we can test whether a node is interpretable by generating questions about the node and checking how accurately we can answer them using only our understanding of the node. In principle, we should be able to perfectly answer any question for a fully interpretable node.

In practice, it is hard to identify questions that are purely about intermediate nodes, for which we can identify ground truth through some other mechanism. Instead, we ask questions about the \emph{entire computation}. Specifically, we construct settings with some latent variable $V$ for which we know the ground truth, and check whether we can predict the value of $V$ using only the information in the node $N$. Note that in this setting, we should no longer expect perfect performance even if $N$ is fully interpretable: it is possible to have some questions about the full computation where the node does not contain the information needed to answer the question.

In principle, these evaluations should be done with humans, to assess interpretability to humans. In practice, these evaluations are typically conducted using LLM autoraters, which serve as a much faster automated proxy for humans.

\paragraph{Monitorability evaluations.} One subset of question answering is of particular interest: \textit{monitorability}. A set of nodes is monitorable for a condition $C$ if, given the input tokens and the values of the intermediate nodes $N$, it is possible to evaluate whether condition $C$ holds of the final response. \citet{guan2025monitoring} develop three types of monitorability evaluations:
\begin{enumerate}
    \item \textbf{``Intervention'' evaluations:} An intervention may or may not be applied to the prompt. Given the chain of thought ($N$) and optionally the response, a classifier must predict whether the intervention has been applied ($V$).
    \item \textbf{``Process'' evaluations:} Suppose an AI system provides the correct answer to a challenging question. Given the chain of thought ($N$) and optionally the response, a classifier must predict what reasoning steps the AI system used to answer the question ($V$).
    \item \textbf{``Outcome property'' evaluations:} Given the chain of thought ($N$) and optionally the prompt, a classifier must predict some property of the response ($V$).
\end{enumerate}

\paragraph{Stress tests.} Since full understanding should allow us to answer \emph{any} question, a more rigorous evaluation is to search for questions that cannot be answered by looking at the intermediate nodes. However, it is then important that the questions are well-chosen such that the information needed to answer them is known to be present in the node. This ensures that if we find a question that we cannot answer, it implies that we are failing to understand the information in the node, providing evidence against interpretability.

\citet{emmons2025chain} and \citet{deng2025cot} provide an example of what such a stress test could look like. They consider settings in which an LLM can only perform well when using chain of thought, and check whether the LLM can evade a monitor while still performing well. In other words, they search for LLM prompts such that the monitor is unable to find the behavior it is looking for from the chain of thought. They find that LLMs struggle with this, providing evidence that the chain of thought is interpretable.

Note that it will usually be theoretically possible for models to evade monitors in such stress tests given sufficient capabilities. For example, for chain of thought monitoring, the Transformer architecture is capable of supporting undetectable steganographic computation~\citep{zamir2024undetectable,christ2024undetectable}, and as such we would expect that fine-tuning the model to do this would eventually work and successfully evade monitors.

\subsection{Human-understandable information}

Intuitively, the key reason that LLM chains of thought are interpretable is because we train LLMs to predict natural language, providing the LLM with a ``natural language prior''. As a result, the model's semantics for the tokens in the chain of thought should mimic the semantics that humans would ascribe to those tokens, because that is what would be useful for predicting text that was generated by humans.

Inspired by this basic argument, we propose two criteria that together suggest that an intermediate node should be considered interpretable:
\begin{enumerate}
    \item \textbf{Natural language prior:} The intermediate representation is directly optimized either (1) to imitate human-written text (e.g., via pre-training or supervised fine-tuning) or (2) to look good to humans (e.g., RLHF~\citep{ziegler2019fine}).
    \item \textbf{Limited optimization towards other incentives:} There are relatively weak training incentives other than towards the natural language prior (ideally none). Incentives that directly push against the natural language prior are worse than incentives that are orthogonal to it.
\end{enumerate}

We illustrate this approach with six examples, in roughly decreasing order of interpretability:
\begin{enumerate}
    \item \textbf{Chain of thought for instruction-tuned language models:} In a language model that has only undergone pre-training and instruction tuning~\citep{ouyang2022training} via supervised fine-tuning and/or RLHF, the chain of thought satisfies both criteria above and thus we expect it to be interpretable.
    \item \textbf{Intermediate steps for text diffusion models:} Typically when training text diffusion models~\citep{li2022diffusion, nie2025large}, a single diffusion step is applied to a noised input, and the resulting output is trained to directly mimic natural language or its token embeddings. Thus, outputs of diffusion steps are directly optimized to imitate natural language, and there is no optimization for other training incentives.\footnote{Note that it is important that only a single diffusion step is applied at training time. If multiple diffusion steps were applied at training time before the resulting output was optimized to imitate natural language, then the intermediate outputs would also be optimized to be useful representations for future diffusion steps.} So, at inference time when multiple diffusion steps are applied, the intermediate outputs at each step should be considered interpretable.
    \item \textbf{Chain of thought for reasoning models:} Reasoning models~\citep{jaech2024openai} undergo reinforcement learning where the rewards are based on successful task completion. This introduces additional incentives: in particular, the chain of thought is now additionally optimized to perform useful intermediate computation, and to be a useful intermediate representation, for completing the assigned task. For current models, we expect that this optimization is substantially weaker than that from pre-training, and so the chain of thought likely remains interpretable. However, substantially scaling RL could make the chain of thought less interpretable.
    \item \textbf{Chain of thought with token superpositions:} Recent work increases the expressivity of inference-time chain of thought by allowing each element of the chain of thought to be the predicted probability distribution over tokens, rather than a single sampled token~\citep{zhang2025soft, zhuang2025text, wu2025llms}. Since the chain of thought is still based on tokens and the training is unchanged, these states should still be considered interpretable. In principle, such an approach could still significantly increase the difficulty of \emph{using} this interpretability, because the state at each point is now exponentially larger. However, \citet{wu2025llms} find that in practice, only a small number of reasoning paths matter, and so the potential exponential blowup does not happen.
    \item \textbf{Continuous chain of thought:} Continuous chain of thought vectors~\citep{hao2024training} may or may not be interpretable. For example, if the vectors passed are simply token embeddings that were optimized via pre-training, they would remain interpretable. However, if they are trained solely to be useful for the model to solve difficult problems, that would likely be uninterpretable.
    \item \textbf{Activations in a Transformer-based language model:} Standard intermediate activations in a Transformer, such as the residual stream or the key vectors for an attention head, are not optimized towards natural language and so do not meet the first criterion. As a result, we should expect that they will not be naturally interpretable.
\end{enumerate}

\paragraph{Evaluation.} In cases where there is optimization towards other incentives, we may not be able to determine a priori whether the resulting nodes will have human-understandable information content or not. In such cases, we can test for this property.

Specifically, we can replace the node's content with paraphrased versions that retain the same (human) semantics, and check the model's resulting performance~\citep{roger2025reasoning}. If the model was only using human-understandable information content, then a semantics-preserving paraphrase should not alter performance. In contrast, if the model relies on node content that is not human-understandable, that content will likely be destroyed by paraphrasing, causing a performance degradation.

In practice, paraphrasing can move a model out of distribution which can degrade performance for innocuous reasons such as spurious correlations, and so \citet{roger2025reasoning} performs light fine-tuning as an initial step for both the original and paraphrased reasoning traces before checking for performance degradation.

Note that unlike question answering evaluations, this evaluation requires that we are able to reverse the translation, that is, given some natural language we can set the intermediate node to have content that corresponds to that natural language. This may not be possible for some architectures, and in those cases we would not be able to run this evaluation.

\paragraph{Caveat: Information content, not reasoning or algorithms.} Note that the natural language prior argument only implies that humans would be able to understand the content of \emph{intermediate values in the overall computation}, whereas it does \textbf{not} imply that we can understand the algorithms or reasoning underlying the computation~\citep{levy2025state}. We can instead try to infer what algorithm or reasoning must have generated those intermediate values.

For LLM chains of thought, the model's reasoning also mimics written human reasoning, and so we can understand both information content and reasoning / algorithms. Text diffusion models are an example where these two properties come apart. If the first intermediate state in a text diffusion model is ``The answer is 8'' while the next state is ``The answer is 7'', we can reasonably conclude that the model changed its best guess for the answer, but we would have to do more work to understand the reasoning that drove the change.

\paragraph{Caveat: Not a guarantee.} Technically, to have a guarantee, the target or loss function should be something that is produced using \emph{only} human semantics ascribed to natural language, thus ensuring that the only information that can be extracted from it is about human semantics. This is a similar property as would be desired in MONA~\citep{farquhar2025mona}, in which non-myopic approval should be provided by a trusted evaluator without looking at external consequences, thus ensuring that any incentives only result from what the evaluator themselves were able to understand.
    
Human-written text and human evaluations do not meet this bar. For example, humans may write about the results of real-world experiments. This text-generation process certainly depends strongly on human semantics and reasoning, but it also depends on real-world physical law. Thus, an AI system trained on this text could in principle infer aspects of physical law that humans may not yet understand.
    
In practice, we expect that such effects are relatively unimportant and can be neglected, and so we treat human-written text and human evaluations as sufficient for imparting a natural language prior that matches human semantics.
Another complication is that this property can be transitive: for example, a loss function that solely encouraged imitation of token embeddings, that themselves were only optimized based on human-written text, would still impart a natural language prior.

\section{By-Hand Depth Calculations for Gemma 3}
\label{appendix:hand-examples}
In this section we provide worked out examples of by-hand depth calculations for the Gemma 3 family of models. These can be quite tedious, so to speed up the process we prompted Gemini 2.5 Pro with the open-source pytorch code of each model along with various instructions to pay attention to important details such as sliding-window size and pre and post-attention norm layers. After some prompt iteration we used human expertise to verify the depth calculations.

\subsection{Gemma 3 1B}

The following is a circuit depth calculation for the full Gemma 3 1B model.

\paragraph{1. Gemma 3 1B Model Architecture Parameters}

We use the relevant architectural parameters for the Gemma 3 1B model from the \href{https://github.com/google-deepmind/gemma/blob/main/gemma/gm/nn/\_gemma.py}{open source code}.

\begin{itemize}
    \item \textbf{\texttt{num\_embed} (V)}: 262,144 (This is the vocabulary size)
    \item \textbf{\texttt{num\_layers}}: 26 (\texttt{\_NUM\_LAYERS\_GEMMA3\_1B})
    \item \textbf{\texttt{embed\_dim} (D)}: 1,152
    \item \textbf{\texttt{hidden\_dim} (Hidden)}: $6 \cdot 1,152 = 6,912$
    \item \textbf{\texttt{num\_heads}}: 4
    \item \textbf{\texttt{head\_dim} (H)}: 256
    \item \textbf{\texttt{use\_post\_attn\_norm}}: True
    \item \textbf{\texttt{use\_post\_ffw\_norm}}: True
    \item \textbf{\texttt{use\_qk\_norm}}: True
    \item \textbf{\texttt{sliding\_window\_size}}: 512
    \item \textbf{Attention Pattern}: A repeating pattern of (5 \texttt{LOCAL\_SLIDING}, 1 \texttt{GLOBAL}). Over 26 layers, this results in \textbf{22 sliding attention layers} and \textbf{4 global attention layers}.
    \item \textbf{Sequence Length (T)}: For sliding attention layers, the effective sequence length is the window size (512). For global attention layers, it is the full sequence length, which we will denote as a variable $T$.
\end{itemize}

For our calculations, we will use the base-2 logarithm ($\log$) and round to the nearest integer for clarity where appropriate.

\begin{itemize}
    \item $\log(V) = \log_2(262144) = \textbf{18}$
    \item $\log(D) = \log_2(1152) \approx 10.16 \rightarrow \textbf{11}$
    \item $\log(\text{Hidden}) = \log_2(6912) \approx 12.75 \rightarrow \textbf{13}$
    \item $\log(H) = \log_2(256) = \textbf{8}$
    \item $\log(\text{sliding\_window\_size}) = \log_2(512) = \textbf{9}$
\end{itemize}

\paragraph{2. Calculating Depth of Core Components}

Following the provided rules, we determine the circuit depth for the fundamental operations and modules. The depth of a component is the length of the longest path from its input to its output.

\begin{itemize}
    \item \textbf{Input Embedding}: This operation looks up a token's vector from an embedding table of size $V$. This is equivalent to a multiplexer, which has a depth of $\log(V)$. It is followed by a scaling multiplication (depth 1).
    \begin{itemize}
        \item \textbf{Depth ($D_{\text{embed}}$) = $\log(V) + 1 = 18 + 1 = 19$}
    \end{itemize}

    \item \textbf{Linear Layer}: A linear transformation $y = Wx$ on a d-dimensional input involves $d$ multiplications computed in parallel (depth 1), followed by a sum of these $d$ results. Using a binary tree of additions, this sum has a depth of $\log(d)$.
    \begin{itemize}
        \item \textbf{Depth = $1 + \log(d_{\text{in}})$}
    \end{itemize}

    \item \textbf{RMSNorm Layer}: This layer computes $x \cdot \text{rsqrt}(\text{mean}(x^2) + \text{eps}) \cdot (1 + \text{scale})$. The longest computational path involves the mean calculation over a vector of dimension $d$.
    \begin{enumerate}
        \item \texttt{square(x)}: 1 (element-wise)
        \item \texttt{sum(square(x))}: $\log(d)$
        \item \texttt{divide\_by\_d}: 1
        \item \texttt{add\_epsilon}: 1
        \item \texttt{rsqrt}: 1 (piecewise function)
        \item \texttt{multiply\_by\_x}: 1
        \item \texttt{multiply\_by\_(1+scale)}: 1
    \end{enumerate}
    \begin{itemize}
        \item \textbf{Depth = $6 + \log(d)$}
    \end{itemize}

    \item \textbf{Attention Mechanism (\texttt{Attention})}: We trace the longest data path through the attention module. The model uses QK Normalization, which adds to the depth.
    \begin{enumerate}
        \item \textbf{Q/K/V Projections} (Linear, D -> H): $1 + \log(D)$
        \item \textbf{QK Norm} (RMSNorm on head dimension $H$): $6 + \log(H)$
        \item \textbf{Rotary Positional Encoding (RoPE)} (element-wise ops): $\approx 2$
        \item \textbf{Query Scaling} (element-wise multiply): 1
        \item \textbf{Q @ K\textsuperscript{T}} (Matrix mul, sum over head dim $H$): $1 + \log(H)$
        \item \textbf{Softmax} (Includes sum over seq length $T$): $2 + \log(T)$
        \item \textbf{Probs @ V} (Matrix mul, sum over seq length $T$): $1 + \log(T)$
        \item \textbf{Output Projection} (Linear, D -> D): $1 + \log(D)$
    \end{enumerate}
    \begin{itemize}
        \item \textbf{Total Attention Depth ($D_{\text{attn}}$)} = $(1+\log D) + (6+\log H) + 2 + 1 + (1+\log H) + (2+\log T) + (1+\log T) + (1+\log D)$
        \item \textbf{$D_{\text{attn}} = 15 + 2 \cdot \log(D) + 2 \cdot \log(H) + 2 \cdot \log(T)$}
    \end{itemize}

    \item \textbf{Feed-Forward Network (\texttt{FeedForward})}: This MLP block consists of two linear projections with a GeLU activation and element-wise multiplication in between.
    \begin{enumerate}
        \item \textbf{Gating Projection} (Linear, D -> Hidden): $1 + \log(D)$
        \item \textbf{GeLU and Multiply} (element-wise): 2
        \item \textbf{Linear Projection} (Linear, Hidden -> D): $1 + \log(\text{Hidden})$
    \end{enumerate}
    \begin{itemize}
        \item \textbf{Total FFW Depth ($D_{\text{mlp}}$)} = $(1+\log D) + 2 + (1+\log(\text{Hidden}))$
        \item \textbf{$D_{\text{mlp}} = 4 + \log(D) + \log(\text{Hidden})$}
    \end{itemize}
\end{itemize}

\paragraph{3. Calculating Depth of a Single Transformer Block (\texttt{Block})}

The longest path through a transformer block traverses all its components sequentially. Gemma 3 uses pre-normalization for both sub-layers and post-normalization for the residual connections (\texttt{use\_post\_attn\_norm=True}, \texttt{use\_post\_ffw\_norm=True}). The critical path is: \texttt{pre\_attn\_norm} -> \texttt{attn} -> \texttt{post\_attn\_norm} -> \texttt{residual\_add} -> \texttt{pre\_ffw\_norm} -> \texttt{mlp} -> \texttt{post\_ffw\_norm} -> \texttt{residual\_add}.

The depth of a single block ($D_{\text{block}}$) is the sum of the depths of these components:
\begin{align*}
    D_{\text{block}} &= D(\text{pre\_attn\_norm}) + D(\text{attn}) + D(\text{post\_attn\_norm}) + D(\text{add}) \\
    &\quad + D(\text{pre\_ffw\_norm}) + D(\text{mlp}) + D(\text{post\_ffw\_norm}) + D(\text{add}) \\
    &= (6+\log D) + (15+2\log D+2\log H+2\log T) + (6+\log D) + 1 \\
    &\quad + (6+\log D) + (4+\log D+\log\text{Hidden}) + (6+\log D) + 1
\end{align*}
\textbf{$D_{\text{block}} = 45 + 7 \cdot \log(D) + 2 \cdot \log(H) + \log(\text{Hidden}) + 2 \cdot \log(T)$}

We now calculate the depth for the two types of attention blocks using our pre-calculated log values.

\begin{itemize}
    \item \textbf{Sliding Block Depth} ($T = 512$, $\log(T) = 9$):
    \begin{align*}
        D_{\text{slide}} &= 45 + 7(11) + 2(8) + 13 + 2(9) \\
        &= 45 + 77 + 16 + 13 + 18 = \textbf{169}
    \end{align*}

    \item \textbf{Global Block Depth} ($T$ is a variable):
    \begin{align*}
        D_{\text{global}} &= 45 + 7(11) + 2(8) + 13 + 2 \cdot \log(T) \\
        &= 45 + 77 + 16 + 13 + 2 \cdot \log(T) = \textbf{151 + $2 \cdot \log(T)$}
    \end{align*}
\end{itemize}

\paragraph{4. Final Calculation for the Entire Gemma 3 1B Model}

The total circuit depth is the sum of the depths of all layers along the longest computational path, from the initial input token to the final logit.

\begin{enumerate}
    \item \textbf{Input Embedding} (Lookup + Scaling):
    \begin{itemize}
        \item $D_{\text{embed}} = \textbf{19}$
    \end{itemize}

    \item \textbf{Transformer Blocks} (22 sliding, 4 global):
    \begin{itemize}
        \item $D_{\text{blocks}} = 22 \cdot D_{\text{slide}} + 4 \cdot D_{\text{global}}$
        \item $D_{\text{blocks}} = 22 \cdot 169 + 4 \cdot (151 + 2 \cdot \log(T))$
        \item $D_{\text{blocks}} = 3718 + 604 + 8 \cdot \log(T) = \textbf{4322 + $8 \cdot \log(T)$}$
    \end{itemize}

    \item \textbf{Final Normalization} (RMSNorm over \texttt{embed\_dim}):
    \begin{itemize}
        \item $D_{\text{final\_norm}} = 6 + \log(D) = 6 + 11 = \textbf{17}$
    \end{itemize}

    \item \textbf{Output Decoding} (Linear Projection to vocab):
    \begin{itemize}
        \item The depth depends on the input dimension (\texttt{embed\_dim}), not the output, as the $V$ dot products are parallel.
        \item $D_{\text{decode}} = 1 + \log(D) = 1 + 11 = \textbf{12}$
    \end{itemize}
\end{enumerate}

\textbf{Total Depth = $D_{\text{embed}} + D_{\text{blocks}} + D_{\text{final\_norm}} + D_{\text{decode}}$}
\begin{align*}
    \text{Total Depth} &= 19 + (4322 + 8 \cdot \log(T)) + 17 + 12
\end{align*}

The final, fully calculated circuit depth for the Gemma 3 1B model is:

\textbf{Depth = $4370 + 8 \cdot \log_2 T$}

where $T$ is the sequence length processed by the four global attention layers. The maximum sequence length is $T=32,768$, so $\log_2 T$ is at most \textbf{15}. Hence the final depth is:
\begin{align*}
    \textbf{Depth} &= 4370 + 8 \cdot \log_2 T = 4370 + 8\cdot15 = 4490
\end{align*}

\subsection{Gemma 3 Family Depth Summary}

We now summarize the total circuit depth calculated for the maximum sequence length (\texttt{T\_max}) of each model in the Gemma 3 family.

First, we calculate the required logarithm values for \texttt{T\_max}:
\begin{itemize}
    \item For Gemma 3 1B: $\log_{2}(\text{T\_max}) = \log_{2}(32,000) \approx 14.97 \rightarrow \mathbf{15}$
    \item For the other models: $\log_{2}(\text{T\_max}) = \log_{2}(128,000) \approx 16.97 \rightarrow \mathbf{17}$
\end{itemize}

\cref{tab:gemma-depth-summary} provides the final depth formulas and the calculated depth at the maximum sequence length for each model.
\cref{tab:gemma-params} shows the key parameters and intermediate calculations that lead to the final depth.

\begin{table}[ht]
\centering
\resizebox{\columnwidth}{!}{%
\begin{tabular}{|l|l|l|l|l|}
\hline
\textbf{Parameter} & \textbf{Gemma 3 1B} & \textbf{Gemma 3 4B} & \textbf{Gemma 3 12B} & \textbf{Gemma 3 27B} \\
\hline
\textbf{num\_layers} & 26 & 34 & 48 & 62 \\
\textbf{embed\_dim (D)} & 1,152 & 2,560 & 3,840 & 5,376 \\
\textbf{hidden\_dim (Hidden)} & 6,912 & 10,240 & 15,360 & 21,504 \\
\textbf{head\_dim (H)} & 256 & 256 & 256 & 128 \\
\textbf{log(D)} & 11 & 12 & 12 & 13 \\
\textbf{log(Hidden)} & 13 & 14 & 14 & 15 \\
\textbf{log(H)} & 8 & 8 & 8 & 7 \\
\textbf{\# Sliding Blocks} & 22 & 29 & 40 & 52 \\
\textbf{\# Global Blocks} & 4 & 5 & 8 & 10 \\
\textbf{Sliding Block Depth} & 169 & 179 & 179 & 185 \\
\textbf{Global Block Depth} & 151 + 2·log(T) & 159 + 2·log(T) & 159 + 2·log(T) & 165 + 2·log(T) \\
\textbf{Total Depth (T=T\_max)} & \textbf{4,490} & \textbf{6,206} & \textbf{8,754} & \textbf{11,662} \\
\hline
\end{tabular}%
}
\caption{Details of the Gemma 3 model intermediate calculations that contribute to the final depth.}
\label{tab:gemma-params}
\end{table}

\end{document}